\documentclass{article}

\usepackage{microtype}
\usepackage{graphicx}
\usepackage{subfigure}
\usepackage{booktabs} 
\usepackage{soul}

\usepackage{hyperref}

\usepackage[accepted]{icml2019}

\icmltitlerunning{Internal representation dynamics and geometry in recurrent neural networks}

\begin{document}

\twocolumn[
\icmltitle{Internal representation dynamics and geometry in recurrent neural networks}



\icmlsetsymbol{equal}{*}

\begin{icmlauthorlist}
\icmlauthor{Stefan Horoi}{udem,mila}
\icmlauthor{Guillaume Lajoie}{udem,mila}
\icmlauthor{Guy Wolf}{udem,mila}
\end{icmlauthorlist}

\icmlaffiliation{udem}{Department of Mathematics and Statistics, Universit\'{e} de Montr\'{e}al;}
\icmlaffiliation{mila}{Mila - Qu\'{e}bec AI institute, Montr\'{e}al, QC, Canada}

\icmlcorrespondingauthor{Guy Wolf}{guy.wolf@umontreal.ca}

\icmlkeywords{Recurrent neural networks, dynamical systems, dimensionality reduction}

\vskip 0.3in
]



\printAffiliationsAndNotice{} 

\begin{abstract}
The efficiency of recurrent neural networks (RNNs) in dealing with sequential data has long been established. However, unlike deep, and convolution networks where we can attribute the recognition of a certain feature to every layer, it is unclear what "sub-task" a single recurrent step or layer accomplishes. Our work seeks to shed light onto how a vanilla RNN implements a simple classification task by analysing the dynamics of the network and the geometric properties of its hidden states. We find that early internal representations are evocative of the real labels of the data but this information is not directly accessible to the output layer. Furthermore the network's dynamics and the sequence length are both critical to correct classifications even when there is no additional task relevant information provided.
\end{abstract}

\section{Introduction}
Research has been done studying the geometry of internal representations and its effect on classification in deep neural networks \cite{Cohen644658}. However, this was not studied in RNNs where recurrent dynamics also play a significant role in the task completion. Therefore, our experiments were chosen with two goals in mind: finding the geometric properties of internal class representations and evaluating the effects of the recurrent dynamics on the classification accuracy.

\section{Methodology and Analysis tools}
We trained a vanilla RNN to complete the well known sequential MNIST classification task, where each input to the network is a sequence of 28 lines (rows) each having 28 pixels. The RNN has a single recurrent layer of 200 tanh neurons and its only parameters are the recurrent and input weights (i.e., with no bias parameters to affect the recurrent dynamics). The output layer is linear and the network is trained using the Adam optimization algorithm with a cross-entropy loss function. After training the network for 30 epochs we achieve an accuracy of a little over 93\% on the test data. While this accuracy is far from the state-of-the-art, it is sufficient to ensure that our network is able to generalize the information provided by the training data to the test data and correctly classify the images in a majority of cases. The code was implemented in Python using PyTorch and all the experiments were conducted using the trained network and the test data, which the network did not see during training.
All experiments reported below are conducted on networks trained in the above manner.
\par
\vspace{-0.05in}
For data analysis, PCA is used to estimate the linear dimensionality of the internal representations of both the entire data as a whole and of the classes. This is done by counting the number of principal components necessary to explain at least 90\% of the variance in the data (or in a class). Also, t-SNE \cite{tsne} is used to create a two dimensional representation of the network's internal states. In this representation, points that are close in the original space will be mapped close together and distant point will be mapped far from one another. The aim is to see how early in the classification process did the network group together similar inputs. Ultimately, we want to see how early does the network "know", in a sense, that a certain input is of a specific class.

\begin{figure*}[ht]
\vskip 0.2in
\begin{center}
\centerline{\includegraphics[width=1.82\columnwidth]{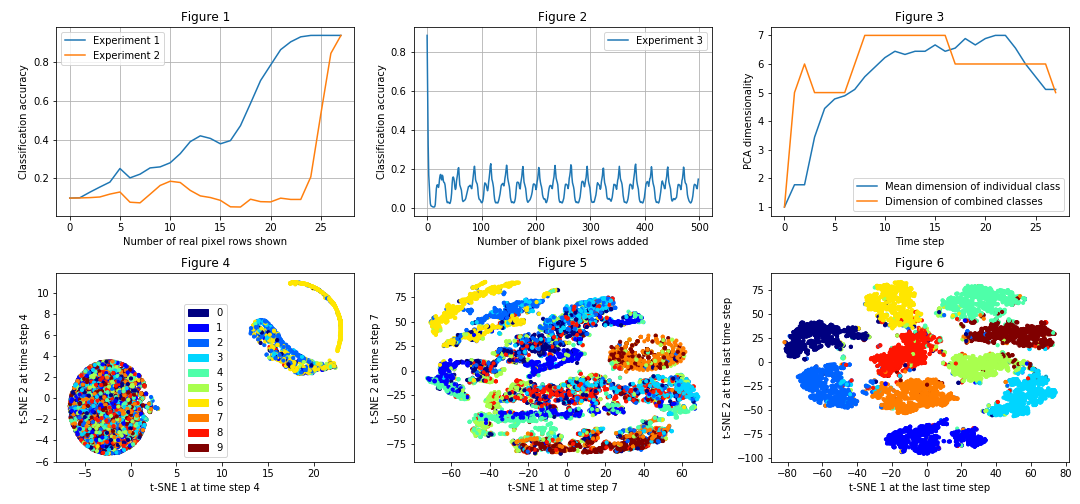}}
\end{center}
\vskip -0.2in
\end{figure*}

\section{Experiments}

\textbf{Experiment 1:} The first experiment consists of modifying the test images so that the last $n$ lines of each image are blank (value 0), this was done for $n$ ranging from 1 to 27. The part-blank images were then used as inputs for the classification task. This experiment aims to determine how the recurrent dynamics alone help the classification task.
\par
\vspace{-0.05in}

\textbf{Experiment 2:} The second experiment consists of giving the network only the first $n$ lines of the images and then stopping, in contrast to Experiment 1. 
This allows to probe the importance of early internal representations by the last linear layer and to see whether or not the network relies on a fixed sequence length for classification.
\par
\vspace{-0.05in}

\textbf{Experiment 3:} In the third experiment, blank pixel rows are added at the end of the full input sequences, increasing their length. This is in order to see how the network's dynamics affect the internal representations after the network was provided with all the available information.
\par
\vspace{-0.05in}

\section{Discussion}
Figure 1a shows that the network's accuracy in experiments 1 and 2 behaves in very different manners. The network dynamics seem to have a significant role in the classification process even when the inputs are blank, as is the case in experiment 1. Despite being shown the same number of real pixel rows, the accuracy in experiment 1 is greatly superior to experiment 2 for all amounts of shown rows. This also indicates that the network's classification abilities are highly dependent on the sequence length even when the amount of relevant information in the sequence is exactly the same.
Figure 1b further emphasizes the importance of the sequence length for proper classification since the accuracy dramatically drops as soon as additional blank rows of pixels are added to the input sequences. The reason we chose to plot the accuracy of the classification for up to 500 added blank rows is to display the highly dynamic nature of the network which seems to begin an oscillatory trajectory as can be deduced from the recurring pattern in the accuracy.


Figure 3 shows that the tendency of neural networks to rely on dimensionality expansions followed by dimensionality reductions in order to perform their tasks which was discussed by \cite{DBLP:journals/corr/abs-1906-00443} and \cite{FUSI201666} seems to be maintained for recurrent neural networks.\par
Finally the t-SNE visualization is especially evocative when the internal representations are colored according to their real digit class. As early as time step 4 (Figure 4) the network seems to create classification relevant clusters in the representation space. In particular, it seems to "know" that certain points are 6es (rightmost cluster) or either 6es or 2s (middle cluster) and it separates them from the rest of the points (leftmost cluster). This internal separation increases in precision throughout the time steps as is shown in Figures 5 and 6, but the initial cluster of "6es" is maintained suggesting that the initial separation of this cluster from the other points was correct.
\section{Conclusion}
Our results show that the network's internal representation is evocative of the real data classification early in the input's sequence so the task is carried out as soon as relevant information is available. Despite the separability of the internal representations and the clusters formed in the representation space, the task relevant information is only interpretable by the output layer after a fixed sequence length. If the sequence length varies the network's dynamics affects the hidden states in a way that greatly hinders accuracy.\par
The retrieval of "early classifications", where the network doesn't wait for the whole input sequence, could be of great impact in time sensitive real-world applications in which a decision is needed as soon as possible. Further work is required to determine how this information is retrievable (see related work \cite{NIPS2017_7188} and \cite{linprobes}). Also, machine learning visualization algorithms show great promise in exposing the inner mechanisms of neural networks and could greatly help in the understanding of these "black-box" algorithms. A great number of vizualisation, dimensionality reduction and manifold learning techniques exist and their applicability in neural network analysis should be evaluated more thoroughly.

\newpage

\onecolumn
\bibliography{example_paper}
\bibliographystyle{icml2019}

\end{document}